\documentclass[10pt,twocolumn,letterpaper]{article}

\usepackage[pagenumbers]{cvpr} 

\usepackage{lipsum}
\usepackage{graphicx}
\usepackage{amsmath}
\usepackage{amssymb}
\usepackage{booktabs}
\usepackage{color}
\usepackage{indentfirst}
\newcommand{\Rui}{\color{black}{}}

\usepackage[pagebackref,breaklinks,colorlinks]{hyperref}

\usepackage[capitalize]{cleveref}
\crefname{section}{Sec.}{Secs.}
\Crefname{section}{Section}{Sections}
\Crefname{table}{Table}{Tables}
\crefname{table}{Tab.}{Tabs.}

\def\cvprPaperID{11} 
\def\confName{CVPR}
\def\confYear{2022}

\begin{document}


\title{What Do Adversarially trained Neural Networks Focus: A Fourier Domain-based Study}

\author{Binxiao Huang,\quad Chaofan Tao,\quad Rui Lin,\quad Ngai Wong\\
The University of Hong Kong\\
huangbx7@connect.hku.hk, \{cftao, rlin, nwong\}@eee.hku.hk
}

\maketitle
\begin{abstract}
Although many fields have witnessed the superior performance brought about by deep learning, the robustness of neural networks remains an open issue. Specifically, a small adversarial perturbation on the input may cause the model to produce a completely different output. Such poor robustness implies many potential hazards, especially in security-critical applications, \textit{e.g.}, autonomous driving and mobile robotics. This work studies what information the adversarially trained model focuses on. Empirically, we notice that the differences between the clean and adversarial data are mainly distributed in the low-frequency region. We then find that an adversarially-trained model is more robust than its naturally-trained counterpart due to the reason that the former pays more attention to learning the dominant information in low-frequency components.
In addition, we consider two common ways to improve model robustness, namely, by data augmentation and by using stronger network architectures, and understand these techniques from a frequency-domain perspective. We are hopeful this work can shed light on the design of more robust neural networks.
\end{abstract}


\section{Introduction}
\label{sec:intro}
Deep neural networks (DNNs) have exhibited strong capability in various domains like image classification~\cite{tao2019minimax}, visual relation detection~\cite{bin2019mr}, motion prediction~\cite{tao2020dynamic} and graph learning~\cite{chen2021litegt}. However, research in adversarial learning shows that even well-trained convolutional neural network (CNN) models are highly susceptible to adversarial perturbations. Those perturbations are nearly indistinguishable to the human eye but can mislead neural networks to completely erroneous outputs, thus endangering safety-critical applications.

Recently, various attack and defense methods have been proposed, and both learn from each other and progress together. PGD attacks are considered as one of the strongest first-order attacks and are commonly used to generate adversarial examples to verify robustness. For the defender, adversarial training (AT), treated as a min-max saddle point problem mathematically, proves to be the most effective defense without obfuscated gradients problems~\cite{athalye2018obfuscated}. The inner-loop maximum problem finds the adversarial examples that are most likely to be misclassified, and the outer-loop optimizes network weights to classify correctly. 
\begin{figure}[t]
    \centering
    \includegraphics[width=0.8\linewidth]{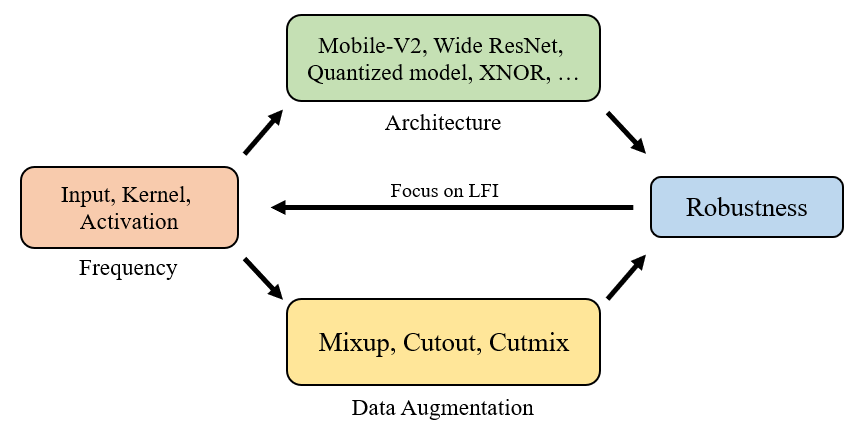}
    \caption{An overview of our Fourier-domain analysis on the robustness of neural networks.}
    \label{fig:overview}
\end{figure}

On the other hand, frequency analysis has been explored to study the nature of neural networks, \textit{e.g.}, model compression~\cite{wang2016cnnpack, lin2022ezcrop}, feature denoising~\cite{zhu2019seismic} and anti-oversmoothing~\cite{wang2022antioversmoothing}. In this work, we are curious about what information the adversarially trained model focuses on. A reason to adopt Fourier-based frequency-domain analysis is that it readily reveals the distribution of information in certain frequencies. This property helps us locate the dominant information in the models and compare the difference between clean and adversarial examples easily.

In practice, we find that the difference between clean and adversarial examples is mainly in the low-frequency region. From the defender's perspective, our frequency-domain analysis and experiments conclude that the robust model focuses more on low-frequency information (LFI) without making much use of high-frequency information (HFI). This indicates that educating the model to utilize HFI can help improve robustness. 

In addition, we consider two common ways, data augmentation, and strong network architecture, to improve robustness and understand the effect of the aforementioned techniques via Fourier analysis. Specifically, three representative data augmentation methods, Mix-up~\cite{zhang2017mixup}, Cutout~\cite{devries2017improved} and  Cutmix~\cite{yun2019cutmix}, are employed in the adversarial training. ResNet~\cite{he2016deep}, EfficientNet~\cite{tan2019efficientnet}, MobileNet-V2~\cite{sandler2018mobilenetv2} and Wide ResNet~\cite{zagoruyko2016wide} are selected as network architectures. We \textit{also} take quantized network architectures into account, since quantization is not only widely used in mobile deployment but also has implications on robustness.


As summarized in Fig.~\ref{fig:overview}, this work novelly adopts a frequency-domain perspective to 1) explain the robustness of a model; 2) interpret data augmentation; and 3) analyze different full-precision and quantized networks.
We empirically find that:
\begin{itemize}
\item The difference between clean and adversarial examples mainly lies in their low-frequency components. The adversarially trained model focuses more on LFI and does not make full use of HFI, and thus is more robust and less accurate than the naturally trained model.


\item A vanilla usage of data augmentation (\textit{e.g.} Mix-up) in adversarial training does not help improve robustness obviously. Mix-up and Cutmix don't explicitly encourage learning low-frequency components and Cutout directly discards some LFI.

\item The robust network architecture with higher similarity between the shallow and deep layers usually has a better accuracy of clean data. Frequency-Aware Transformation~\cite{tao2021fat} smooths the network weights and thus improves the robustness of models, especially for quantized models.



\end{itemize}

\section{Related Work}
\label{sec:lit_review}

{\Rui{This section provides an overview of popular adversarial attacks and defensive approaches. In addition, we take particular interest in the works studying the robustness of the quantized neural networks (QNNs), which are lightweight for edge devices but suffer robustness issues under attacks.}}

{\Rui{\subsection{Attack Methods}}}
\label{subsec:attack}

Neural networks are susceptible to adversarial perturbations, first discovered in~\cite{szegedy2013intriguing}. Fast Gradient Sign Method (FGSM)~\cite{goodfellow2014explaining} was proposed to generate adversarial examples using first-order gradients. Projected Gradient Descent (PGD)~\cite{madry2017towards} iterates FGSM several times to strengthen the attacks. Along this line, different variants were proposed, e.g., Basic Iterative Method (BIM)~\cite{kurakin2016adversarial} and Carlini and Wagner (CW)~\cite{carlini2017towards}. Ref.~\cite{madry2017towards} proved that models that are robust to PGD should be robust to all first-order attacks. Therefore, PGD is regarded as a standard method for evaluating the robustness of networks in practice.

{\Rui{\subsection{Defensive Methods}}}
\label{subsec:defensive}
{\Rui{
Adversarial Training (AT) is proved to be a standard defensive method to generate robust models~\cite{madry2017towards}. Some AT methods focus on the generation of proper adversarial examples~\cite{shafahi2019adversarial, cai2018curriculum, zhang2020attacks}, some try to control the training by adding more constraints to the network itself~\cite{kannan2018adversarial, mao2019metric, wang2019improving}. However, those approaches cannot convincingly explain the large gap between robustness and accuracy. Ref.~\cite{ilyas2019adversarial} argues that the understanding of humans and networks about the input information is different. It is echoed by~\cite{wang2020high}, claiming networks can sense HFI that is almost imperceptible to humans, and smooth kernels are preferred to robust models, which motivates us to zoom in deeper robustness analysis from a frequency viewpoint.

{\subsection{Data Augmentation}}
Data augmentation aims to prevent over-fitting problems and improve the generalization capability of the networks, which is essential to robust networks. Therefore, we conduct extensive experiments to explore whether typical data augmentation methods~\cite{zhang2017mixup,devries2017improved,yun2019cutmix} can boost robustness. The proposed mix-up~\cite{zhang2017mixup} performs a weighted average of selected pairs of examples within the same batch to get a new one. In Cutout~\cite{devries2017improved}, the pixels in a randomly chosen patch are zeroed when given an image. While in Cutmix~\cite{yun2019cutmix}, the chosen patch is replaced by its counterpart from another randomly selected image.

}}

{\Rui{\subsection{Robustness of QNNs}}}
\label{subsec:robust_quant}
{\Rui{
QNNs can preserve comparable accuracy to their full-precision counterparts, but their robustness lags a long way behind due to their inferior representation ability. To alleviate the robustness issue, Ref.~\cite{chmiel2020robust} employs a variety of quantization scenarios with a Kurtosis regularization (KURE) on the weights. Ref.~\cite{lin2019defensive} suppresses the distance between the outputs of the clean and adversarial images, adding a regularizer based on the Lipschitz constant of the network. In~\cite{tao2021fat}, an attention module is introduced, which filters out the quantization-unfriendly parts of the weights during the training of QNNs in the frequency-domain.
}}

\section{Analyses}
\label{sec:exp}

 We take ResNet-20 as a backbone network and use SGD optimizer with a momentum of 0.9 and a global weight decay of $5\times10^{-4}$ in experiments. The model is trained for 120 epochs with a batch size of 512 on two 3090 GPUs. The cross-entropy (CE) loss function is employed for the classification task. The initial learning rate is 0.1, decayed by a factor of 0.1, 0.1, 0.5 at 60, 90, and 110 epochs, respectively. All experiments are performed on the CIFAR-10 dataset, which contains 50k training examples and 10k test examples. Because we do not know the mean and standard deviation of the attack examples in the real world, the training examples are cropped to $3\times32\times32$ size to be fed into the model without normalization.

 Following~\cite{madry2017towards}, we use PGD-7 adversarial training as a standard method and check the robustness against various attacks. Echoing the training, the robustness of the PGD-20 attack equipped with random-start is taken as the main basis for robustness analysis since it proves to be one of the strongest first-order attacks. The attack step is  $\alpha = 2/255$ and the constraint is $\epsilon=8/255$.

\subsection{frequency-domain Analysis}
\label{sec:frequency}
The visualization of the frequency-domain shows that the difference between clean and adversarial examples is mainly distributed in the low-frequency region. The input analysis proves that the LFI is important to adversarially trained models. Compared with naturally trained models, the kernels of adversarially trained models are smoother and have fewer bright spots in the high-frequency region. Besides, for the activations of the first layer that reflect the information extracted directly from the image, the LFI proportion increases significantly during adversarial training. Therefore, adversarial training makes the model more focused on LFI, allowing it to classify adversarial examples better to improve robustness.

\subsubsection{Frequency Analysis of Input}
\label{sec:f_input}
\begin{figure}[t]
  \centering
  %
    \includegraphics[width=0.8\linewidth]{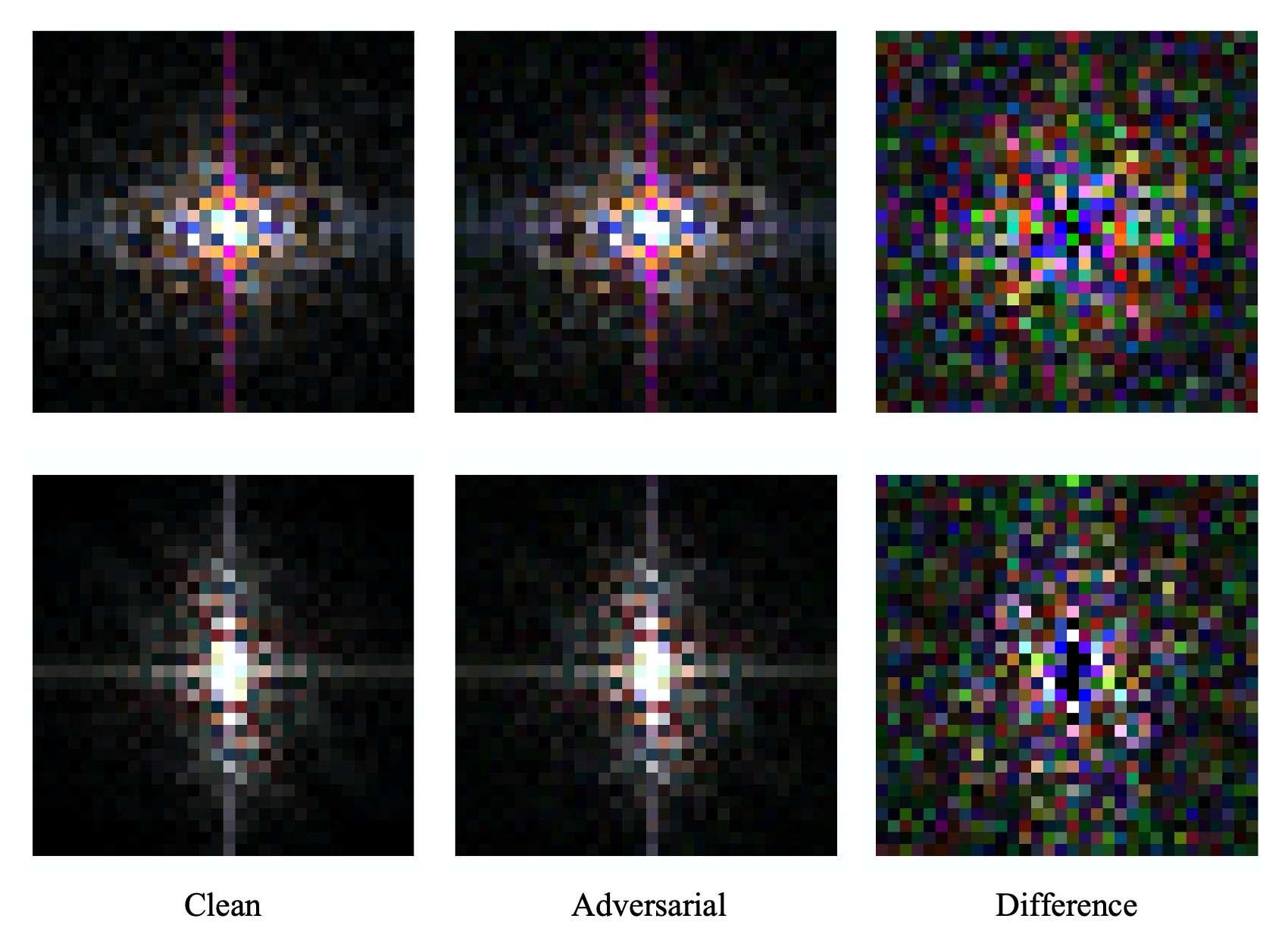}
   \caption{Visualization of clean data (left), adversarial examples (middle) generated by the PGD-20 attack, and the difference (right) after DFT and normalization in the frequency-domain, with low frequencies in the center and high frequencies around. The brighter the pixel, the higher the frequency amplitude. The difference is mainly concentrated in the low-frequency region.}
   \label{fig1}
\end{figure}

Adversarial examples are usually generated by adding crafted negligible perturbations to natural examples. The intuitive idea is to perform a frequency analysis to see if the introduced perturbations are mainly concentrated in the high-frequency region. Fig.~\ref{fig1} shows the frequency views of clean and adversarial examples by discrete Fourier transform (DFT) based on an adversarially trained ResNet-20 model on the CIFAR-10 dataset.

As shown in Fig.~\ref{fig1}, the difference is mainly distributed in the low-frequency region, with smaller amplitudes in the high-frequency region. The adversarial examples need to satisfy the $l_p$ norm constraints. Low-pass filtering (LPF) will retain a large amount of LFI, so the picture will not change too much and can meet the requirements. 

As for defense, the attack is mainly concentrated in the low-frequency region, but there is still some in the high-frequency region. 
We apply different levels of LPF to see the robustness of multiple attacks and the accuracy of the clean data. Degree 16 means that after DFT, only the $16\times16$ patch in the center of the picture is saved, and all external values are zeroed in the frequency-domain. As Table~\ref{tab1} shows, the loss of HFI leads to a negligible decrease in clean data accuracy, while the robustness of several attacks is improved slightly, indicating that misleading information plays a more significant role than useful information in high-frequency domains for the adversarially trained model. The clean data accuracy (76.67\%) is similar to the naturally trained network (77.14\%), only using low-frequency components in~\cite{wang2020high}. So the results show that the adversarially trained models focus more on LFI and do not take full advantage of HFI. 

\setlength{\tabcolsep}{0.7mm}
\begin{table}
\footnotesize
\caption{Top-1(\%) accuracy of ResNet-20 trained with PGD-7 attack on the CIFAR-10 dataset against various attacks ($\epsilon=8/255$). The number in the LPF column represents the degree of LPF. The higher the value, the more information is retained. 32 means all information is reserved. The 20 after the attack represents 20 iterations, and the same is true for the later table.}

\label{tab1}
\centering
\scriptsize
\setlength{\tabcolsep}{0.9mm}{
\begin{tabular}{l|c|c|c|c|c|c|c|c}
\toprule
Model & LPF & Natural & GN & FGSM & PGD-20 & BIM-20 & TPGD-20 & CW-20 \\
\hline
ResNet-20 & 32 & \textbf{76.67} & 71.24 & 49.34 & 43.12 & 43.01 & 62.45 & 39.78\\

ResNet-20 & 28 & 76.60 & 72.38 & 49.56 & 43.76 & 43.64 & 62.85 & 61.06\\

ResNet-20 & 24 & 76.28 & 73.34 & 49.52 & 44.22 & 44.12 & 63.27 & 64.69\\

ResNet-20 & 20 & 75.70 & \textbf{73.90} & 49.57 & 44.99 & 44.92 & 64.03 & \textbf{64.80}\\

ResNet-20 & 16 & 74.18 & 73.00 & \textbf{49.67} & \textbf{46.12} & \textbf{45.98} & \textbf{64.53} & 63.83\\

\bottomrule
\end{tabular}}
\end{table}

Why cannot adversarial training learn HFI well? It is mentioned in~\cite{wang2020high} that neural network training first learns LFI in the early stage and focuses on learning HFI in the later stage for a fixed dataset. But during the adversarial training, the input adversarial examples are changed in each step. It is understandable that the model cannot capture the HFI in a single cycle to improve the performance further. Besides, we conducted a comparison experiment with different batch sizes. The smaller the batch size, the more frequently the adversarial examples are updated, and the model focuses less on HFI, leading to a slight increase in adversarial robustness. Of course, it is not true that the smaller the batch size, the more robust the model is. If the update frequency is too fast, resulting in poor learning of HFI, it will reduce both the accuracy and robustness. This experiment has better robustness and accuracy with a batch size of around 128. The results are shown in Table~\ref{bs}.

In conclusion, the difference between clean and adversarial examples is mainly distributed in the low-frequency region, and adversarially trained models focus more on LFI.

\setlength{\tabcolsep}{0.7mm}
\begin{table}
\footnotesize

\caption{Top-1(\%) accuracy of ResNet-20 trained with different batch size ($\epsilon=8/255$).}

\label{bs}
\centering
\scriptsize
\setlength{\tabcolsep}{0.6mm}{
\begin{tabular}{l|c|c|c|c|c|c|c|c}
\toprule
Model & Batch size & Natural & GN & FGSM & PGD-20 & BIM-20 & TPGD-20 & CW-20 \\
\hline
ResNet-20 & 512 & 75.02 & 70.70 & 48.09 & 42.89 & 42.87 & 62.40 & 40.75\\

ResNet-20 & 256 & 76.38 & 70.67 & 49.43 & 43.11 & 43.03 & 62.30 & 39.08\\

ResNet-20 & 128 & \textbf{76.89} & 71.33 & \textbf{50.14} & \textbf{44.31} & \textbf{44.23} & \textbf{63.60} & 40.58\\

ResNet-20 & 64  & 75.97 & \textbf{71.54} & 49.51 & 43.71 & 43.72 & 62.86 & \textbf{40.86}\\

ResNet-20 & 32  & 74.92 & 69.97 & 47.05 & 41.71 & 41.80 & 61.77 & 40.05\\

\bottomrule
\end{tabular}}
\end{table}

\subsubsection{Frequency Analysis of Kernel}
\label{f_kernel}
\begin{figure}[t]
  \centering
    \includegraphics[width=0.8\linewidth]{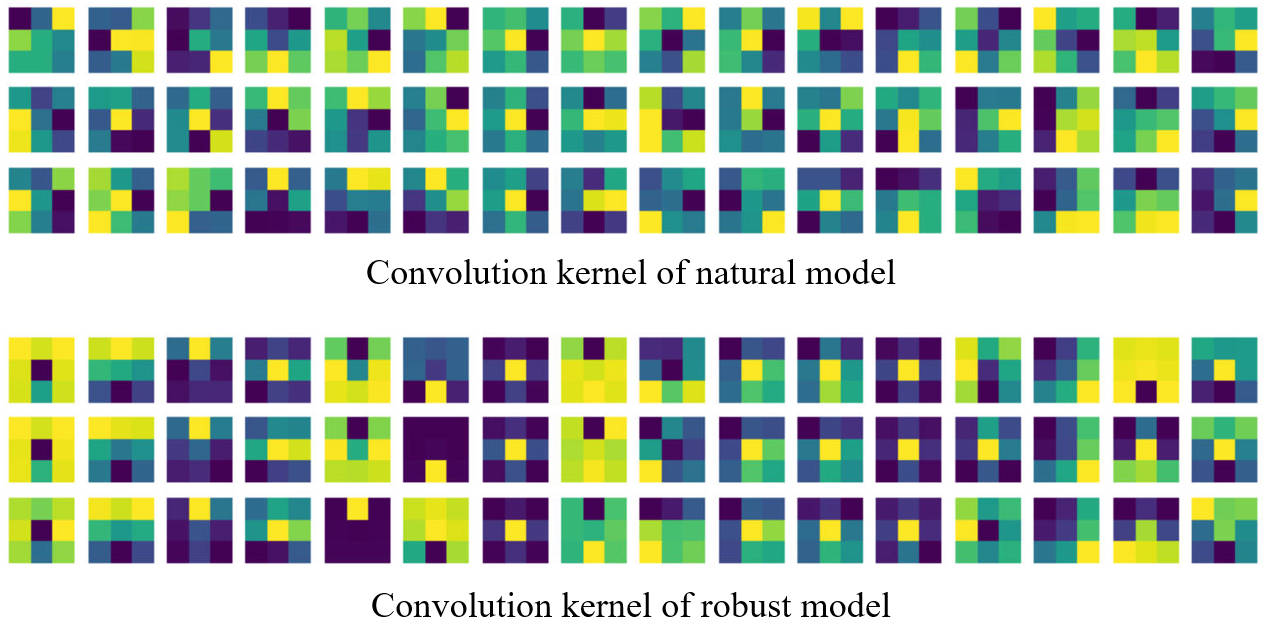}
   \caption{Visualization of first convolution's kernels of a naturally trained model (top) and an adversarially trained model (bottom). 3 channels vertically and 16 kernels horizontally. Robust model has smoother convolution kernels than natural model.}
   \label{fig3}
\end{figure}

To explore the difference between different robustness models, we obtained a naturally trained model and an adversarially trained model. Fig.~\ref{fig3} illustrates their first convolution weights. The convolution has 16 kernels, each kernel has 3 channels, and the size is $3\times 3$. The weights of the robust model have a similar distribution in three channels, and the distribution is smoother than the natural model. 

\begin{figure}[t]
  \centering
    \includegraphics[width=0.8\linewidth]{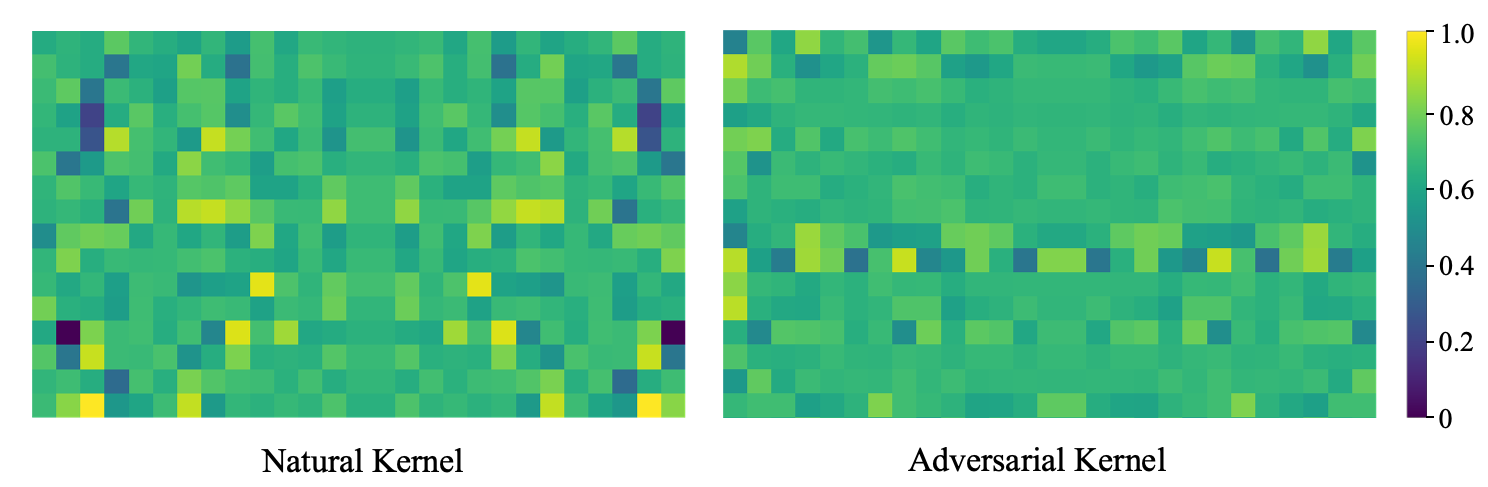}
   \caption{Frequency maps of first convolution kernels in a naturally trained model (left) and an adversarially trained model (right). Kernels in robust model have fewer bright spots in the high-frequency region.}
   \label{fig4}
\end{figure}

 In addition, we adjust the size of first convolution weight of each layer in ResNet-20 from $[c_{out}, c_{in}, H, W]$ to $[c_{out}, c_{in} * H * W]$, and perform DFT on the last dimension. In Fig.~\ref{fig4}, the two ends of the graph are low-frequency regions, and the middle is the high-frequency region. The convolution operation in the spatial domain is equivalent to the element-wise multiplication in the frequency-domain. The robust model has more bright spots in the low-frequency region and fewer bright spots in the high-frequency region, which means that the robust model kernels pay more attention to the less-frequency signals. So adversarially trained networks are more capable of recognizing LFI and do not learn well in high-frequency regions. Combined with the previous observation that the main difference between clean and adversarial examples is in the low-frequency region shown in Fig.~\ref{fig1}, this can also explain why the adversarially trained models are robust.

\subsubsection{Frequency Analysis of Activation}
\label{activation}
The activation of the first layer dealing directly with the inputs reflects the information that the network extracts from the inputs. Therefore, we investigate the ratio of LFI and HFI in the first layer's activation to further verify whether adversarially trained models focus more on LFI. For ResNet-20 model, $Activation = hardtanh(bn1(conv1(input)))$. Ref.~\cite{wang2022antioversmoothing} refers to LPF as a filter that only preserves the Direct-Current (DC) component, while diminishing the remaining high-frequency component. The LFI (DC component) and HFI in a matrix $X$ could be calculated as follows without running DFT explicitly: $LFI[x] = \frac{1}{n}\textbf{1}\textbf{1}^T x $, $HFI[x] = (\textbf{I} - \frac{1}{n}\textbf{1}\textbf{1}^T) x $.
The ratio of LFI to HFI is defined as $\textbf{R} = \frac{\left\|LFI[x]\right\|_2}{\left\|HFI[x]\right\|_2}$. 

\begin{figure}[t]
  \centering
    \includegraphics[width=0.8\linewidth]{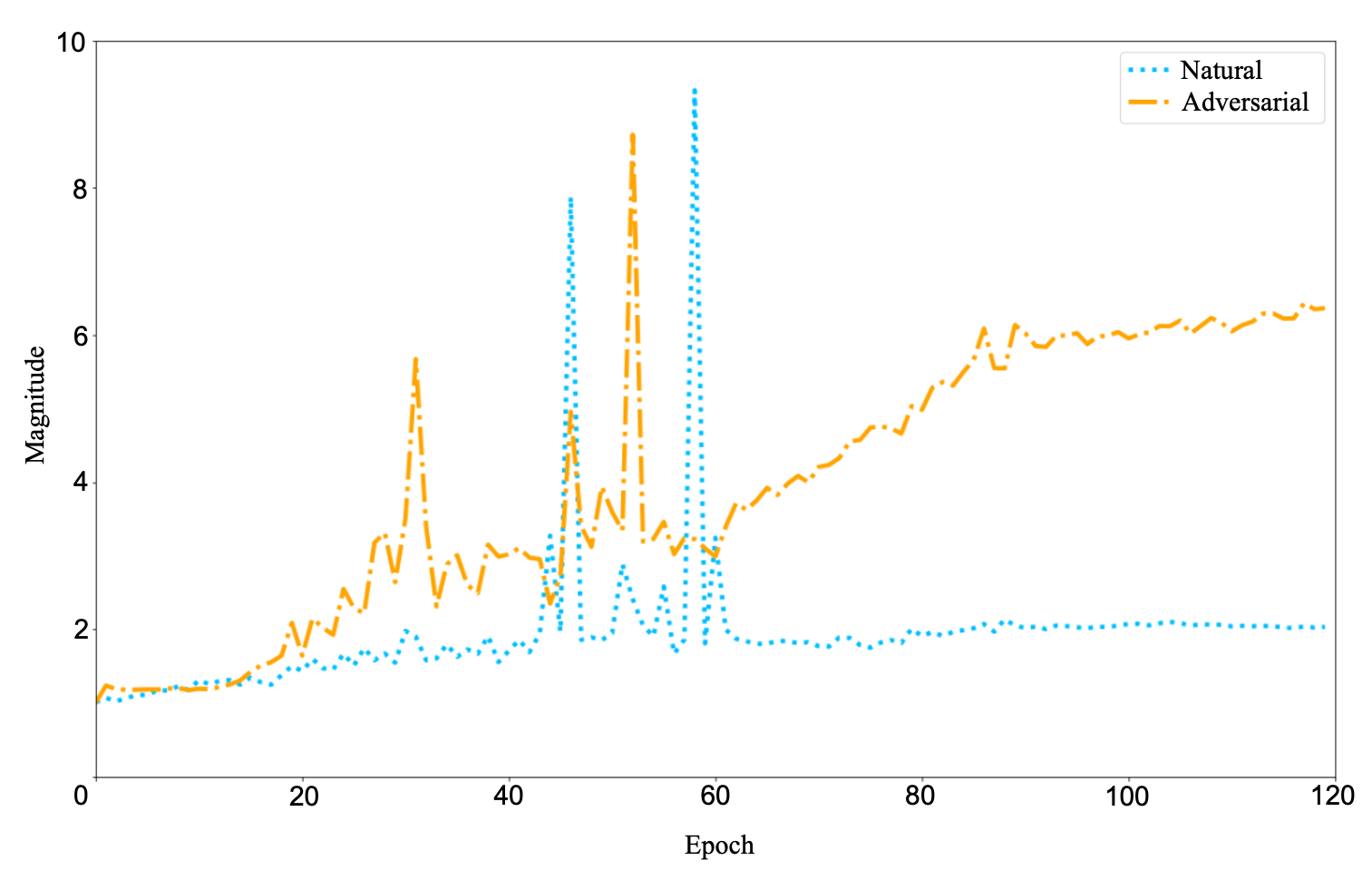}
   \caption{The ratio of LFI to HFI. Although some individual epochs change dramatically, the adversarial model gradually increases the focus on LFI.}
   \label{lp_hp}
\end{figure}

$\textbf{1}\in\mathbb{R}^n, x\in\mathbb{R}^{n\times d}$ means n-length d-channel signals, $\textbf{I}$ means an unit matrix, and $T$ means transpose operation. These equations have been mathematically proved in \cite{wang2022antioversmoothing}. Within each training epoch, we no longer use shuffle to break up the data and take the first batch examples to calculate the ratio. We reshape the activation from $[B, C, H, W]$ to $[B, C, H * W]$, and use the previously mentioned equations to calculate the ratio of LFI to HFI of the last two dimensions, then get an average ratio in the first dimension. The magnitudes of $\textbf{R}$ in natural training and adversarial training are shown in Fig.~\ref{lp_hp}. The $\textbf{R}$ of the natural model improves from 1.01 to 2.03, while the adversarial model improves from 1.00 to 6.36. 

Frequency analysis of activations reveals that the adversarially trained models pay more attention to LFI.

\begin{figure}[t]
  \centering
    \includegraphics[width=0.8\linewidth]{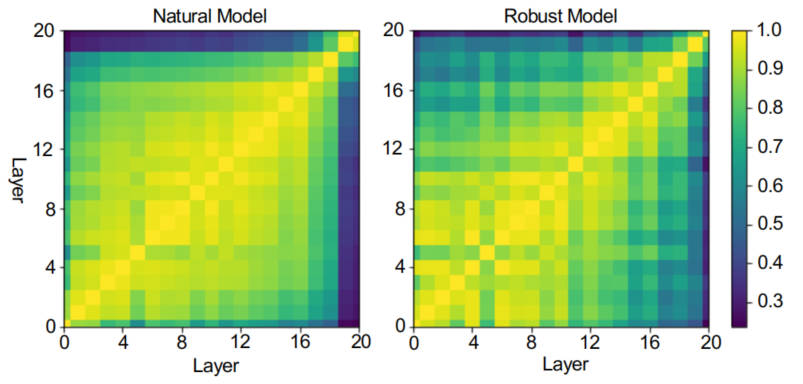}
   \caption{Linear CKA reveals consistent relationships of activations between layers of CNNs. The left side is naturally trained model, and the right side is an adversarially trained model. Robust model has a lower similarity between shallow and deep layers.}
   \label{fig5}
\end{figure}

We also investigate the similarity of activations between shallow and deep layers in natural and robust models by the Centered Kernel Alignment (CKA)~\cite{kornblith2019similarity} approach. The PGD-20 attack examples are fed into the models to obtain the activations of convolution or full connection layer. As shown in Fig.~\ref{fig5}, the value of CKA$[i,j]$ represents the similarity calculated by linear CKA between $i-th$ and $j-th$ activations in the network, and the higher the value, the higher the similarity. The natural model has a high similarity between shallow and deep layers, while the robust model has a low similarity. This means that the natural model could attend to higher-order information in the first few layers of the network, while the adversarial model focuses more on the lower-order aspects. The inability of the robust network to focus on higher-order information upfront may account for the low accuracy of the clean data.

 in Summary, adversarially trained models pay more attention to LFI and have low similarity between shallow and deep layers.

\subsection{Effect of Data Augmentation}

Data augmentation is a common way to improve the generalization ability and robustness of models. Clean data are fed into the attacked model to generate adversarial examples. We could either mix up first and then generate the adversarial examples, or first generate the adversarial examples and then mix up. Some methods will use two examples to generate a new one, so the ground truth labels need to be mixed up as well, and the loss function will change from CE loss to the binary cross-entropy (BCE) loss. We consider 3 representative data augmentation techniques for robustness improvement as follow:

\begin{itemize}
\item Mix-up\cite{zhang2017mixup}: We randomly select a example within the same batch for a weighted average with all the pixels of the original example. Each example will be used only once. Mathematically, $x^i_{k} = \alpha \cdot x^i_k + (1-\alpha) \cdot x^j_k$. The value of $\alpha$ follows the beta distribution, $\alpha \in \left[0, 1\right]$. The subscript $k$ can be $c$ or $a$, representing clean and adversarial examples, respectively. The superscripts $i$ and $j$ represent the indexes within a batch.

\item Cutout\cite{devries2017improved}: We randomly set the value of a patch of the image to 0. The size of the excised area in proportion to the original image conforms to the beta distribution. For an image, $x^i_{k}[:, h_0: h_0+\Delta h, w_0:w_0+\Delta w] = 0$.

\item Cutmix\cite{yun2019cutmix}: We randomly select a patch of the image and replace it with another patch from the same location of another image. The proportion of the area of the patch to the original image is in line with the beta distribution. $x^i_{k}[:, h_0: h_0+\Delta h, w_0:w_0+\Delta w] = x^j_{k}[:, h_0: h_0+\Delta h, w_0:w_0+\Delta w]$.

\end{itemize}


\setlength{\tabcolsep}{0.7mm}
\begin{table}
\footnotesize

\caption{Robustness of different Data Augmentation(DA) methods on CIFAR-10 against various attacks ($\epsilon=8/255$). Here A-mixup means mix-up on adversarial examples, and C-mixup means mix-up on clean data.}

\label{fp-mixup}
\centering
\scriptsize
\setlength{\tabcolsep}{1.1mm}{
\begin{tabular}{l|c|c|c|c|c|c|c}
\toprule
 DA &Natural & GN & FGSM   & PGD-20   & BIM-20& TPGD-20 & CW-20  \\

\hline

 None       & \textbf{74.91} & \textbf{70.44} & \textbf{47.93} & \textbf{42.13} & \textbf{42.15} & \textbf{62.32} & 40.64 \\
 
 A-mixup    & 73.98 & 69.73 & 45.65 & 39.87 & 39.90 & 58.46 & 40.31 \\

 C-mixup    & 62.66 & 61.10 & 43.23 & 40.89 & 40.87 & 56.11 & 46.29 \\
 
 A-cutout   & 69.72 & 68.07 & 35.92 & 27.61 & 27.56 & 52.88 & 28.46 \\

 C-cutout   & 70.10 & 66.04 & 41.86 & 36.49 & 36.46 & 57.00 & 40.84 \\

 A-cutmix   & 72.15 & 67.17 & 45.84 & 40.61 & 40.55 & 59.38 & 44.14 \\

 C-cutmix   & 62.63 & 59.37 & 42.95 & 40.72 & 40.75 & 56.59 & \textbf{48.35} \\
 
\bottomrule
\end{tabular}}
\end{table}

By analyzing the results of Table~\ref{fp-mixup}, it can be concluded that data augmentation on clean data is more robust than applying it to adversarial examples, but the accuracy of clean data degrades. Mix-up and Cutmix have similar performance, but Cutout performed poorly. Mix-up and Cutmix encourage the model to focus more on the high-frequency components and do not encourage anything about lower-frequency components explicitly~\cite{wang2020high}. While cutout forces the network to attend on less discriminative parts, it also cuts out some LFI. Since adversarially trained models rely primarily on LFI to improve robustness, the direct usage of these methods cannot improve the robustness.

\subsection{Effect of Network Architecture}

\subsubsection{Different Full-precision Architectures}
A variety of commonly used classical convolution neural networks, such as EfficientNet\cite{tan2019efficientnet} , MobileNet-V2\cite{sandler2018mobilenetv2}, Wide ResNet\cite{zagoruyko2016wide}, are trained using the standard PGD-7 training method, and the statistical results on CIFAR-10 dataset are shown in the Table~\ref{tab3}. 

\setlength{\tabcolsep}{0.7mm}
\begin{table}
\footnotesize

\caption{Top-1(\%) accuracy of different models trained by PGD-7 attacks on CIFAR-10 against various attacks ($\epsilon=8/255$).}

\label{tab3}
\centering
\scriptsize
\setlength{\tabcolsep}{1.1mm}{
\begin{tabular}{l|c|c|c|c|c|c|c}
\toprule
 model &Natural & GN & FGSM   & PGD-20   & BIM-20& TPGD-20 & CW-20  \\

\hline

 ResNet-20       & 76.67 & 71.24 & 49.34 & 43.12 & 43.01 & 62.45 & 39.78\\

 MobileNet-V2       & 82.97 & 75.07 & 53.41 & 44.42 & 44.44 & 65.25 & 30.24 \\

 EfficientNet   & 76.10 & 71.64 & 49.02 & \textbf{45.41} & 45.40 & 63.91 & \textbf{46.39} \\

 WRN-34-10      & \textbf{85.85} & \textbf{78.54} & \textbf{54.94} & 44.45 & \textbf{45.60} & \textbf{68.47} & 26.58 \\

\bottomrule
\end{tabular}}
\end{table}

\begin{figure}[t]
  \centering
    \includegraphics[width=0.8\linewidth]{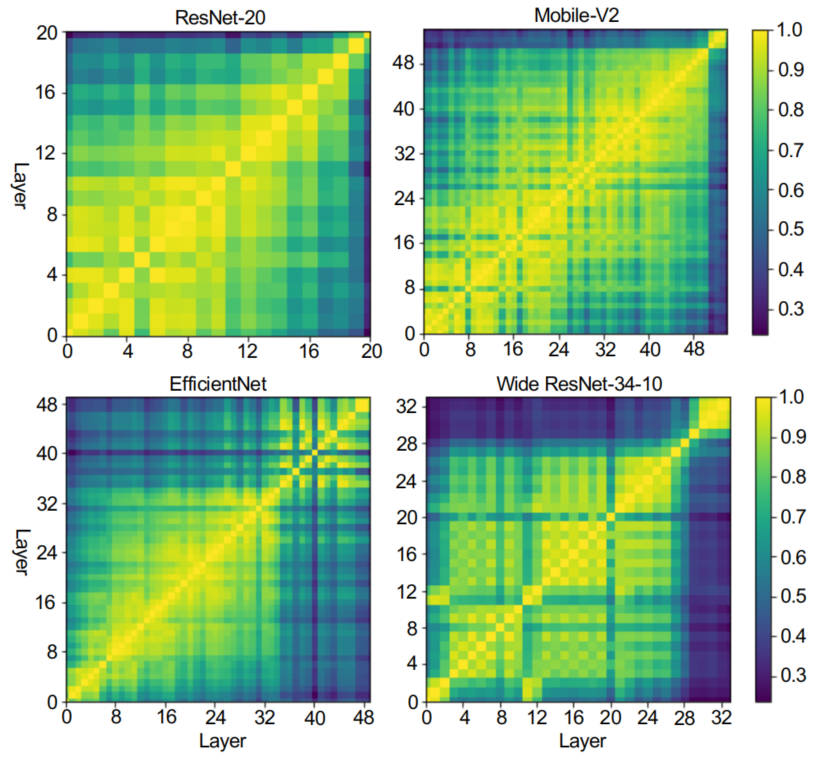}
   \caption{Linear CKA visualization between layers of different architectures on CIFAR-10 dataset. Models with high similarity between the shallow and deep layers have high accuracy on the clean data.}
   \label{arch}
\end{figure}

Taking ResNet-20 (43.12\%) as the baseline against PGD-20 attack, EffientNet (45.41\%), MobileNet-V2 (44.42\%) and Wide ResNet-34-10 (44.45\%) all show better robustness because of the larger capacity. But the accuracy of clean data does not improve by the same margin. We analyze it using Linear CKA, shown in Fig.~\ref{arch}. The robust models with higher accuracy of the clean data, such as MobileNet-V2 (82.97\%), Wide ResNet-34-10 (85.85\%), have a high similarity between shallow and deep layers. While the robust models with lower accuracy, such as ResNet-20 (76.67\%), EfficientNet (76.10\%), have a low similarity between shallow and deep layers. This observation aligns with the discussion in~\ref{activation}.


\subsubsection{Different Low-precision Architectures}

It is generally believed that the larger the capacity, the better the robustness of the model. However, it is difficult to deploy high-performance full-precision networks on devices with limited computational resources. In this case, QNNs are appealing for their reduced storage need and efficient inference. How to ensure the robustness of the quantized network with the limited capacity is a problem that needs to be solved.

Inspired by Ref.~\cite{tao2021fat}, we use a Frequency-Aware Transformation (FAT) module to learn and analyze the kernels in the frequency-domain. The model is replaced with a quantized model, and the rest of the settings are consistent with the standard adversarial training. QM-n means the model is uniformly quantized to n-bit. The robustness of quantized models with or without the FAT module is shown in Table~\ref{tab5}. The experimental results show a small robustness improvement (0.37\%) for the full-precision network with the addition of the FAT block against PGD-20. There is a big gap of robustness between the quantized and full-precision networks, and the adjustment of kernels in the frequency-domain effectively reduces this gap. For 4-bit and 2-bit quantized models, the robustness improves by 3.72\% and 7.77\%, respectively. 

\begin{figure}[t]
  \centering
    \includegraphics[width=0.8\linewidth]{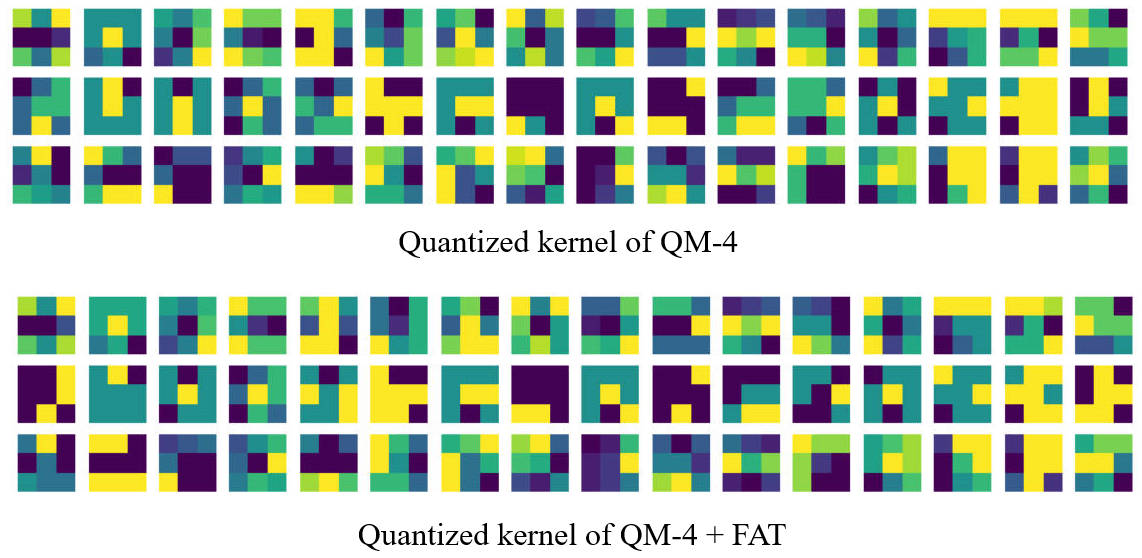}
   \caption{Visualization of quantized kernels with (bottom) or without (top) FAT module. FAT could soften the kernels, which in turn improves the robustness of the quantized model.}
   \label{w2}
\end{figure}

 We illustrate the first quantized convolution weights (not the first convolution, which is not quantized) of the QM-4 model and QM-4 + FAT model in Fig.~\ref{w2}. The adaptive learning module in the frequency-domain can assist in smoothing the weights of the network.

\setlength{\tabcolsep}{0.7mm}
\begin{table}
\footnotesize

\caption{Top-1(\%) accuracy of models trained by PGD-7 attacks on CIFAR-10 against various attacks ($\epsilon=8/255$). QM-n means that the model is compressed to n bit, n = 32 means the full precision model. QM-n + FAT means that an adaptive mask treatment is applied to the weights of QM-n model in frequency-domain.}

\label{tab5}
\centering
\scriptsize
\setlength{\tabcolsep}{1mm}{
\begin{tabular}{l|c|c|c|c|c|c|c}
\toprule
 model  &Natural & GN & FGSM   & PGD-20   & BIM-20& TPGD-20 & CW-20  \\

\hline
 QM-32          & \textbf{76.67} & \textbf{71.24} & \textbf{49.34} & 43.12 & 43.01 & \textbf{62.45} & 39.78 \\

 QM-32 + FAT    & 74.84 & 69.41 & 48.49 & \textbf{43.49} & \textbf{43.47} & 62.40 & 42.25 \\

 QM-4           & 67.58 & 65.13 & 42.16 & 38.75 & 38.50 & 56.81 & \textbf{43.81} \\

 QM-4 + FAT     & 74.58 & 68.98 & 48.11 & 42.47 & 42.80 & 61.92 & 38.82 \\
 
 QM-2           & 57.10 & 54.96 & 34.92 & 32.68 & 32.70 & 50.51 & 38.52 \\
 
 QM-2 + FAT     & 70.37 & 66.79 & 45.90 & 40.45 & 39.99 & 58.95 & 39.25 \\

\bottomrule
\end{tabular}}
\end{table}

\section{Conclusion}
\label{sec:conclusion}
In this paper, we investigate what information the adversarially trained model focuses on via Fourier domain analysis. Studies of inputs, kernels, and activation indicate that adversarially trained networks focus more on the LFI of images and learn HFI poorly. We also find that models with high similarity between shallow and deep layers usually have high accuracy of clean data. In addition, We explain the failure of vanilla usages of data augmentation to improve robustness from a frequency-domain perspective. Our experiments also demonstrate that applying the FAT module to the kernels could improve robustness, especially for QNNs.

\clearpage
{\small
\bibliographystyle{ieee_fullname}
\bibliography{robustness}
}

\end{document}